\documentclass{article}

\usepackage{arxiv}

\usepackage[utf8]{inputenc} 
\usepackage[T1]{fontenc}    
\usepackage{hyperref}       
\usepackage{url}            
\usepackage{booktabs}       
\usepackage{amsfonts}       
\usepackage{nicefrac}       
\usepackage{microtype}      
\usepackage{lipsum}
\usepackage{graphicx}
\usepackage{natbib}
\usepackage{amsmath}

\usepackage{multirow}
\usepackage{amsthm}

\newtheorem{remark}{Remark}

\title{Coupled Inference in Diffusion Models for Semantic Decomposition}

\author{
 Calvin Yeung$^*$, Ali Zakeri, Zhuowen Zou, Mohsen Imani \\
  Department of Computer Science\\
  University of California, Irvine\\
  Irvine, CA 92697 \\
  $^*$Corresponding author: \texttt{chyeung2@uci.edu} \\
}

\begin{document}
\maketitle
\begin{abstract}
Many visual scenes can be described as compositions of latent factors. Effective recognition, reasoning, and editing often require not only forming such compositional representations, but also solving the decomposition problem. One popular choice for constructing these representations is through the binding operation. Resonator networks, which can be understood as coupled Hopfield networks, were proposed as a way to perform decomposition on such bound representations. Recent works have shown notable similarities between Hopfield networks and diffusion models. Motivated by these observations, we introduce a framework for semantic decomposition using coupled inference in diffusion models. Our method frames semantic decomposition as an inverse problem and couples the diffusion processes using a reconstruction-driven guidance term that encourages the composition of factor estimates to match the bound vector. We also introduce a novel iterative sampling scheme that improves the performance of our model. Finally, we show that attention-based resonator networks are a special case of our framework. Empirically, we demonstrate that our coupled inference framework outperforms resonator networks across a range of synthetic semantic decomposition tasks.
\end{abstract}    
\section{Introduction}

Many visual scenes can be described as compositions of latent factors: objects and attributes, entities and relations, parts and wholes. Effective recognition, reasoning, and editing often require not only forming such compositional representations, but also solving the decomposition problem: given a representation, how can we recover the underlying semantic constituents? Recent object-centric approaches exemplify this need by learning to split scenes into object-level slots or components, demonstrating improvements in structured understanding and manipulation, such as Slot Attention~\citep{locatello2020object} and MONet~\citep{burgess2019monet}. Yet, reliably inferring constituents from a single composite embedding remains challenging in general, often involving combinatorial search over factor choices, which quickly becomes intractable as the number of objects and attributes grow. 

One popular choice for constructing compositional representations is through the binding operation \citep{kleykoSurveyHyperdimensionalComputing2023}. In this framework, attributes, e.g., $\textbf{red}$, $\textbf{apple}$, are assigned to distributed representations \citep{gayler1998multiplicative, kanerva2009hyperdimensional,plate}. 
An object can then be represented by combining these constituent factors through the binding operation, e.g., $\textbf{red}\odot\textbf{apple}$, which is often implemented as element-wise multiplication. This way, the object is represented by the conjunction of all its attributes. This way of constructing representations enables the explicit encoding of structure, which is useful for tasks such as visual reasoning \citep{herscheNeurovectorsymbolicArchitectureSolving2023} and object-centric learning \citep{object-centric-learning}.  

\begin{figure*}[!t]
    \centering
    \includegraphics[width=\textwidth]{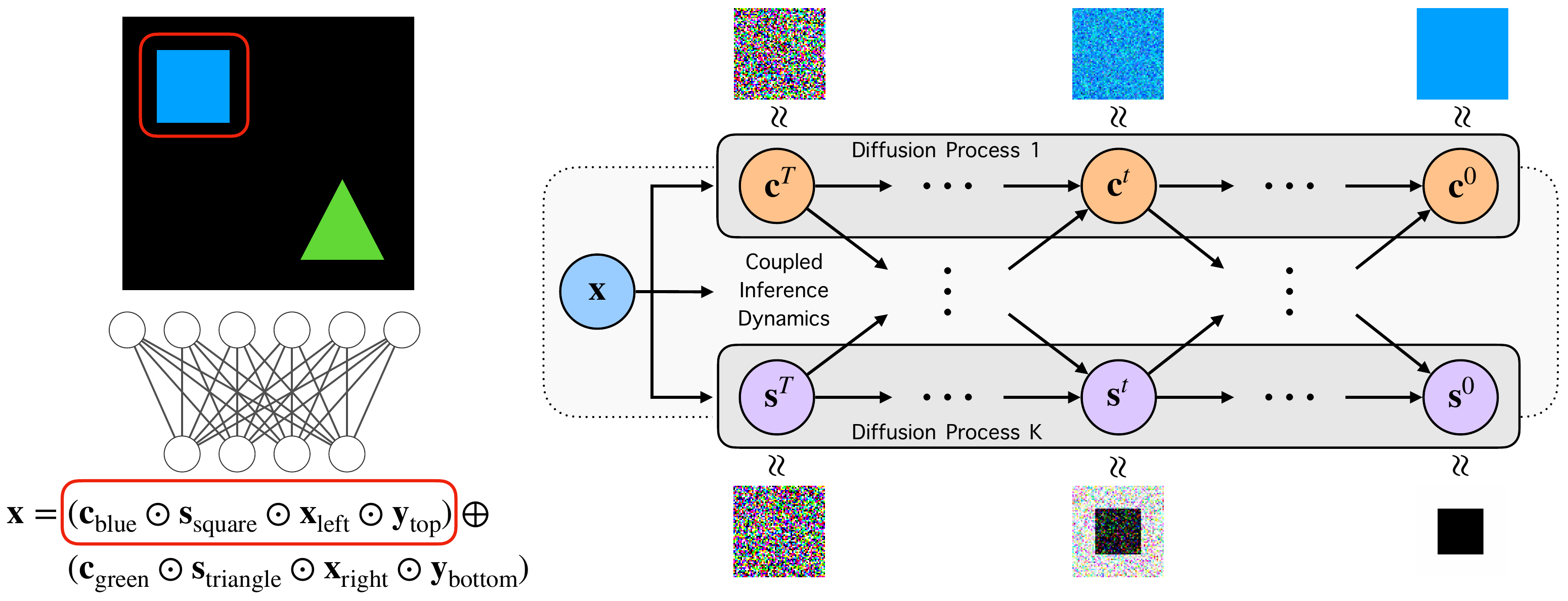}
    \caption{Overview of the semantic decomposition task and methods. \textbf{Left:} Images are passed through a neural network to generate binding-based compositional representations. We would like to identify each constituent using the codebooks that encode their possible values. \textbf{Right:} Coupled diffusion processes for semantic decomposition. Diffusion processes corresponding to each factor operates in parallel and are jointly guided to converge towards the correct factorization.}
    \label{fig:main-fig}
\end{figure*}

Resonator networks were proposed as a way to perform decomposition on such representations. Rather than exhaustively enumerating factor combinations, a resonator network performs coupled recurrent updates over $K$ factor estimates, each constrained to lie in its respective codebook~\citep{fradyResonatorNetworksEfficient2020,kentResonatorNetworks22020}. These dynamics can be understood as $K$ coupled Hopfield networks \citep{hopfieldNeuralNetworksPhysical1982a}, where each codebook defines attractors and the recurrent interactions enforce consistency with the observed composite vector. 

Recent works have shown notable similarities between Hopfield networks and diffusion models \citep{ambrogioniSearchDispersedMemories2024,phamMemorizationGeneralizationEmergence2025}. Since resonator networks are coupled Hopfield networks, a natural question is whether diffusion models can be coupled together to perform semantic decomposition. 

We approach this question by framing semantic decomposition as an inverse problem, where the goal is to infer the constituent factors given the bound representation. 
This is directly analogous to classical imaging and vision problems, e.g., denoising, deblurring and inpainting, where one inverts a forward corruption model using structural priors~\citep{darasSurveyDiffusionModels2024}. Recent advances show that diffusion models are powerful generic priors for such inverse problems: once trained, they can be combined with task-specific guidance to solve the problem without retraining. By assigning each factor in the decomposition problem with a corresponding diffusion prior, we can couple their inference processes through joint guidance.

In this work, we introduce a framework for semantic decomposition using coupled inference in diffusion models. We construct analytic diffusion priors over each factor space, yielding closed-form score functions. By defining an additional energy term, we enforce consistency between the factors and the bound vector, and perform joint diffusion over all factors guided by this energy. Conceptually, our method frames semantic decomposition as a continuous-time coupled denoising process: factor estimates evolve together under the sum of (i) prior scores that keep them near plausible codebook-induced modes, and (ii) a reconstruction-driven guidance term that encourages their composition to match the bound vector. We also introduce a novel iterative sampling scheme that improves the performance of our model. Finally, we show that attention-based resonator networks are a special case of our framework. Empirically, we demonstrate that our coupled inference framework outperforms resonator networks across a range of synthetic semantic decomposition tasks.
\section{Background}
\subsection{The Decomposition Problem}
The decomposition problem is the problem of recovering the constituent vectors of a composite vector generated through the binding operation. Each factor has a discrete set of possibilities (e.g., red, green, blue), represented by vectors stored in a corresponding codebook. 

Suppose we have $K$ codebooks, each with $n$ bipolar vectors of dimension $D$, i.e., $ X_1,\dots, X_K\in\{-1,1\}^{D\times n}$ where each entry is sampled from $\mathcal{U}\{-1,1\}$. This can be generalized beyond bipolar vectors, e.g., vectors in $\mathbb{R}^D$. We write $\mathbf{x}_j\in  X_j$ if $\mathbf{x}_j$ is a member of codebook $X_j$. Let $\mathbf{x}_j\in  X_j$ be some choice of vector from the $j$-th codebook, for $j=1,\dots,K$. We can obtain the composite vector through binding (i.e. element-wise multiplication) $\mathbf{x}=\mathbf{x}_1\odot\cdots\odot\mathbf{x}_K\in\{-1,1\}^D$.
Given the composite vector $\mathbf{x}$ and codebooks $ X_1,\dots, X_K$, we would like to recover the factors $\mathbf{x}_{1:K}$. The search space of this problem is exponential in the number of factors, i.e., $n^K$. 

\subsection{Resonator Networks}\label{sec:resonator}
Resonator networks were introduced as an approximate method for solving the decomposition problem \citep{fradyResonatorNetworksEfficient2020,kentResonatorNetworks22020,rennerNeuromorphicVisualScene2024}. The main idea behind the resonator network is to recurrently apply an update rule on estimates $\hat{\mathbf{x}}^{(1)},\dots,\hat{\mathbf{x}}^{(K)}$:
\begin{align}\label{eq:resonator-update}
\hat{\mathbf{x}}^{(j)}_{t+1} &= \mathrm{sgn}( X_j X_j^\top(\mathbf{x}\odot \mathbf{r}_t^{(j)})),\quad\text{for }j=1,\dots,K
\end{align}
where $\mathbf{r}_t^{(j)}=\bigodot_{i\neq j}\mathbf{x}^{(i)}_t$ and $\mathbf{x}^{(j)}_0=\frac{1}{n}\sum_{\mathbf{x}\in X_j}\mathbf{x}$ for $j=1,\dots,K$.

\paragraph{Relation to Hopfield networks}\label{sec:hopfield}
Hopfield networks are auto-associative memories where the vectors in the codebook $X$ form fixed-point attractors in the dynamics defined by the Hopfield update rule, provided that the number of memorized vectors is below the capacity of network \citep{hopfieldNeuralNetworksPhysical1982a}. A Hopfield network minimizes the energy function $E(\xi)=-\frac{1}{2}\mathbf{x}^\top(XX^\top)\mathbf{x}$
via a dynamical update rule
\begin{align}\label{eq:update1}
    \mathbf{x}_{t+1}=\mathrm{sgn}(XX^\top \mathbf{x}_t)
\end{align}
The network has a certain capacity; if there are too many stored vectors or if the stored vectors are highly correlated, the attractors might not correspond to codebook vectors. 

If each codebook vector has elements sampled i.i.d from $\mathcal{U}(\{-1,1\})$, it has been shown that the memory capacity of network is approximately $0.138D$, where $D$ is the dimension of the vector \citep{PhysRevLett.55.1530}. Modern variants have improved capacity exponential in $D$ \citep{krotov,ramsauerHopfieldNetworksAll2021}.

Thus, \textit{resonator networks can be seen as $K$ coupled Hopfield networks}, where the $j$-th network has the codebook vectors $\mathbf{x}\in  X_j$ as fixed-point attractors. Under the assumption each Hopfield network is not over capacity, the correct solution to the decomposition problem $\hat{\mathbf{x}}^{(j)}=\mathbf{x}_j$ for $j=1,\dots,K$ is a fixed point in the dynamical system defined by the resonator network update rule.

\subsection{Diffusion Models}\label{sec:diffusion}
A diffusion model is a generative model that maps Gaussian noise to data. Given a distribution $p_*(\mathbf{x}_0)$, a forward diffusion process can be used to convert this distribution into Gaussian noise. This process can be expressed as an SDE of the form
\begin{align}
    d\mathbf{x}_t=f(\mathbf{x}_t,t)dt+g(t)d\mathbf{W}_t
\end{align}
where $\mathbf{x}_0\sim p_*$, $t\in[0,1]$ and $f(\mathbf{x}_t,t)$ and $g(t)$ are drift and diffusion terms respectively \citep{songScoreBasedGenerativeModeling2021}. The forward SDE has a corresponding reverse SDE with the same marginals at each time $t$ \citep{ANDERSON1982313}
\begin{align}
    d\mathbf{x}_t=\Big(f(\mathbf{x}_t,t)-g^2(t)\nabla_{\mathbf{x}_t}\log p(\mathbf{x}_t)\Big)dt+g(t)d\bar{\mathbf{W}}_t
\end{align}
which allows us to sample $\mathbf{x}_0\sim p_*$. This reverse SDE requires knowledge of the score function $\nabla_{\mathbf{x}_t}\log p(\mathbf{x}_t)$. To estimate the score using a neural network $\boldsymbol{s}_\theta(\mathbf{x}_t,t)$, we can optimize a conditional score-matching loss
\begin{align}
    \mathcal{L}(\theta)
    =\mathbb E
      \Big[\big\|\boldsymbol{s}_\theta(\mathbf{x}_t,t)-\nabla_{\mathbf{x}_t}\log p(\mathbf{x}_t|\mathbf{x}_0)\big\|^2\Big]
\end{align}
where the expectation is over $t\sim\mathcal{U}([0,1]),\mathbf{x}_0\sim p_*,\mathbf{x}_t\sim p(\mathbf{x}_t|\mathbf{x}_0)$ and the conditional score $\nabla_{\mathbf{x}_t}\log p(\mathbf{x}_t|\mathbf{x}_0)$ is available in closed form.

\paragraph{Probability-flow ODE}
The reverse SDE has an associated probability-flow ODE with the same marginals:
\begin{align}
    \frac{d\mathbf{x}_t}{dt}=f(\mathbf{x}_t,t)-\frac{1}{2} g^2(t)\,\boldsymbol{s}_\theta(\mathbf{x}_t,t).
\end{align}
This yields deterministic sampling and enables tractable likelihood estimation via change of variables. In practice, discrete-time Denoising Diffusion Probabilistic Models \citep{ddpm} correspond to an Euler-Maruyama discretization of a variance-preserving SDE with $f(\mathbf{x},t)=-\frac{1}{2} \beta(t)\,\mathbf{x}$ and $g(t)=\sqrt{\beta(t)}$, while deterministic samplers (e.g., Denoising Diffusion Implicit Models \citep{songDenoisingDiffusionImplicit2022}) correspond to integrating this ODE.

\paragraph{Relation to Hopfield Networks}
There are notable connections between diffusion models and associative memories \citep{ambrogioniSearchDispersedMemories2024,phamMemorizationGeneralizationEmergence2025}. It can be shown that the energy function of a diffusion model at a fixed diffusion step $t$ is equivalent to the log-sum-exp energy function of a modern Hopfield network with a softmax-based update rule \citep{ambrogioniSearchDispersedMemories2024,ramsauerHopfieldNetworksAll2021}. However, there are some differences: (1) while Hopfield networks have deterministic dynamics, diffusion models typically add an additional stochastic term; (2) Hopfield energy is constant over time, while the corresponding score function for diffusion models is time dependent.

\subsection{Inverse Problems}
Inverse problems are problems where we would like to recover $\mathbf{x}$ given $\mathbf{y}$ given a corruption model
\begin{align}
    \mathbf{y}=A(\mathbf{x})+\sigma_y\mathbf{z},\quad \mathbf{z}\sim\mathcal{N}(0,I)
\end{align}
and that $\mathbf{x}\sim p_x$ \citep{darasSurveyDiffusionModels2024}, i.e., we would like to find $p(\mathbf{x}|\mathbf{y})$. $A$ is called the corruption operator. For example, denoising corresponds to when $A=I$. 

Notably, many methods have recently been proposed that aim to solve inverse problems by modeling $p_X$ as a diffusion model. These methods can be characterized as trying to sample from the distribution $p(\mathbf{x}|\mathbf{y})$ by estimating the conditional score
\begin{align}
    \nabla_{\mathbf{x}_t}\log p(\mathbf{x}_t|\mathbf{y})=\nabla_{\mathbf{x}_t}\log p(\mathbf{x}_t)+\nabla_{\mathbf{x}_t}\log p(\mathbf{y}|\mathbf{x}_t)
\end{align}
The unconditional score $\nabla_{\mathbf{x}_t}\log p(\mathbf{x}_t)$ is already given by the diffusion model, while the measurements matching term $\nabla_{\mathbf{x}_t}\log p(\mathbf{y}|\mathbf{x}_t)$ is in general intractable, as the conditional likelihood is usually only defined at $t=0$. Thus, exact computation would require integration over all time steps from $t$ to $0$. Proposed solutions differ in their estimation of this term. We refer the reader to \cite{darasSurveyDiffusionModels2024} for a comprehensive survey.
\section{Methods}
Our method can be motivated through two different perspectives: (1) as a diffusion-based solution to an inverse problem; and (2) as the diffusion model equivalent of resonator networks. We follow the first perspective in this section and discuss the second in section \ref{sec:relation-to-rn}.

\subsection{Decomposition as an Inverse Problem}\label{sec:decomp-inv}
We can frame decomposition as recovering $\mathbf{x}_{1:K}$ given $\mathbf{x}$ when
\begin{align}
    \mathbf{x}=A(\mathbf{x}_{1:K})=\mathbf{x}_1\odot\dots\odot\mathbf{x}_K
\end{align}
where $\mathbf{x}_j\sim p_j=\frac{1}{n}\sum_{\mathbf{x}\in X_j}\mathcal{N}(\mathbf{x}_j;\mathbf{x},\sigma_0^2 I)$ for $j=1,\dots,K$, which we can model as diffusion models. In particular, when $\sigma_0^2\to 0$, each prior $p_j$ becomes a mixture of Dirac-deltas centered at codebook vectors in $ X_j$, exactly recovering the decomposition problem in section \ref{sec:resonator}. Thus, this formulation can be thought of as a decomposition problem with relaxed constraints. Ideally, we would like $\sigma_0^2= 0$ but in practice take $\sigma_0^2>0$ for numerical stability.

An alternate formulation can be expressed as recovering $\mathbf{z}_{1:K}$ given $\mathbf{x}$ when 
\begin{align}\label{eq:latent-decomp}
    \mathbf{x}= X_1\mathbf{z}_1\odot\dots\odot  X_K\mathbf{z}_K
\end{align}
where $\mathbf{z}_j\sim \frac{1}{n}\sum_{i=1}^n \mathcal{N}(\mathbf{z}_j;\mathbf{e}_i,\sigma_0^2 I)$ for $j=1,\dots,K$, where $\mathbf{e}_i\in\mathbb{R}^n$ is the $i$-th one-hot vector. $\mathbf{z}_j$ are latent representations of the codebooks vectors in $X_j$, so $p_j$, when modeled as a diffusion model, is a latent diffusion model \citep{rombachHighResolutionImageSynthesis2022}.

The methodology corresponding to both formulations are very similar. Below, we focus on the former, but we will analyze both in the results section. See section \ref{sec:latent-decomp} in the appendix for more information.

\subsection{Analytical Construction of the Score Function}
As we explicitly specify the form of the data distribution, we can derive a closed form for the score function. Suppose we have a diffusion process over $\mathbf{x}_j^t$ for $t\in [0,1]$ such that  
\begin{align}
    p_j(\mathbf{x}_j^0)&=\frac{1}{n}\sum_{\mathbf{x}\in X_j}\mathcal{N}(\mathbf{x}_j^0;\mathbf{x},\sigma_0^2 I)\\
    p_j(\mathbf{x}_j^1)&=\mathcal{N}(0,I)
\end{align}
We can interpolate between these two distributions through a Gaussian probability path \citep{holderriethIntroductionFlowMatching2025} and express the marginals at time $t$ as
\begin{align}\label{eq:marginal}
    p_j(\mathbf{x}_j^t)=\frac{1}{n}\sum_{\mathbf{x}\in X_j}\mathcal{N}(\mathbf{x}_j^t;\alpha_t\mathbf{x},\omega_t^2I)
\end{align}
where $\omega^2_t=\alpha_t^2\sigma_0^2 +\beta_t^2$ and $\alpha_t,\beta_t:[0,1]\to \mathbb{R}_{\geq 0}$ are continuously differentiable monotonic functions such that $\alpha_0=\beta_1=1$ and $\alpha_1=\beta_0=0$. 

Given the closed form of the marginal, we can get an analytical form for the score function
\begin{align}\label{eq:uncond-score}
    \nabla_{\mathbf{x}_j^t} \log p_j(\mathbf{x}_j^t) &= \frac{1}{\omega^2_t}\sum_{\mathbf{x}\in X_j}\gamma_\mathbf{x}(\mathbf{x}_j^t,t)(\alpha_t\mathbf{x}-\mathbf{x}_j^t)\\
    \text{where}\quad\gamma_\mathbf{x}(\mathbf{x}_j^t,t) &= \frac{\mathcal{N}(\mathbf{x}_j^t;\alpha_t\mathbf{x},\omega_t^2I)}{\sum_{\mathbf{x}'\in X_j} \mathcal{N}(\mathbf{x}_j^t;\alpha_t\mathbf{x}',\omega_t^2I)}
\end{align}
In addition, Tweedie's formula gives us an a denoised estimate given the sample at time $t$ \citep{tweedie}:
\begin{align}
    \hat{\mathbf{x}}_j^0(t) &= \mathbb{E}[\mathbf{x}_j^0|\mathbf{x}_j^t]=\frac{1}{\alpha_t}(\mathbf{x}_j^t+\beta_t^2\nabla_{\mathbf{x}_j^t}\log p_j(\mathbf{x}_j^t))\\
    &=\frac{\beta_t^2}{\omega_t^2}\sum_{\mathbf{x}\in X_j}\gamma_\mathbf{x}(\mathbf{x}_j^t,t)\mathbf{x}+\frac{\alpha_t\sigma_0^2}{\omega_t^2}\mathbf{x}_j^t\label{eq:tweedie}
\end{align}
\begin{remark}\label{rem:softmax}
    Consider the special case where $\sigma_0^2=0$ and codebooks vectors $\mathbf{x}_j\in X_j$ have the same magnitude. Then
\begin{align}
    \hat{\mathbf{x}}_j^0(t)&= X_j\mathrm{softmax}\left(\frac{\alpha_t}{\beta_t^2} X_j^\top \mathbf{x}_j^t\right)
\end{align}
so the denoised estimate at time $t$ is equivalent to a single application of the softmax update rule of a modern Hopfield network \citep{ramsauerHopfieldNetworksAll2021}.
\end{remark}

\subsection{Joint Inference for Semantic Decomposition}\label{sec:joint-inf}
Following the characterization of semantic decomposition as an inverse problem in section \ref{sec:decomp-inv}, we would like to sample from the posterior
\begin{align}
    p(\mathbf{x}_{1:K}|\mathbf{x})\propto p(\mathbf{x}|\mathbf{x}_{1:K})\prod_{j=1}^K p_j(\mathbf{x}_j)
\end{align}
Note that $p_j(\mathbf{x}_j)$ for $j=1,\dots,K$ are independent as they are defined by the corresponding codebooks $ X_j$ which are generated by sampling i.i.d. vectors. 

In practice, we do not require an explicit normalized likelihood $p(\mathbf{x} | \mathbf{x}_{1:K})$. Instead, following diffusion posterior sampling (DPS) \citep{chungDiffusionPosteriorSampling2024}, we guide the reverse diffusion process using gradients of a task-specific energy function evaluated at denoised estimates. Specifically, at diffusion time $t$, the conditional score is approximated as
\begin{align}\label{eq:dps-approx}
    \nabla_{\mathbf{x}_j^t} \log p(\mathbf{x}_{1:K}^t | \mathbf{x})
\approx \nabla_{\mathbf{x}_j^t} \log p_j(\mathbf{x}_j^t) + \nabla_{\mathbf{x}_j^t} \mathcal{E}(\hat{\mathbf{x}}_{1:K}^0(t); \mathbf{x})
\end{align}
where $\hat{\mathbf{x}}_j^0(t)$ is the Tweedie denoised estimate of $\mathbf{x}_j$ at time $t$ (Eq.~\ref{eq:tweedie}), and $\mathcal{E}$ is an energy function encoding our reconstruction objective.

\begin{figure*}[!t]
    \centering
    \includegraphics[width=\textwidth]{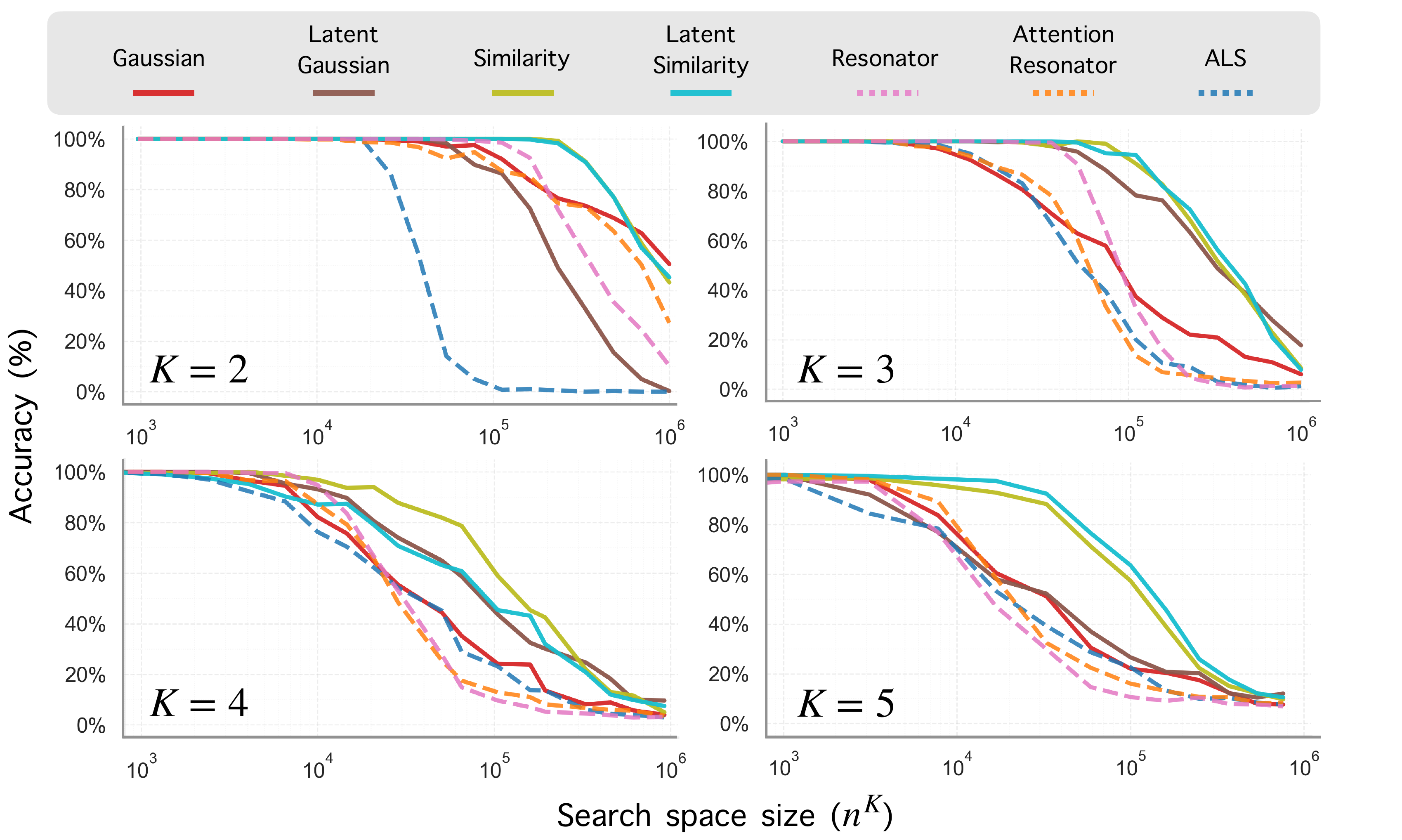}
    \caption{Decomposition accuracy for varying search space sizes. Codebook vectors have dimension $D=1000$ and models are run for 100 iterations.}
    \label{fig:acc-vs-search-space}
\end{figure*}

\paragraph{Gaussian energy}  
If we model the energy as $\log p(\mathbf{x}|\mathbf{x}_{1:K})$ where $p(\mathbf{x}|\mathbf{x}_{1:K})=\mathcal{N}\left(\bigodot_{i=1}^K \mathbf{x}_i,\eta^2 I\right)$, this reduces to a least-squares objective
\begin{align}
    \mathcal{E}_{\text{Gauss}}(\mathbf{x}_{1:K}; \mathbf{x}) = -\frac{1}{2\eta^2} \left\| \mathbf{x} - \bigodot_{i=1}^K \mathbf{x}_i \right\|^2
\end{align}
which gives us the gradient
\begin{align}
    \nabla_{\mathbf{x}_j} \mathcal{E}_{\text{Gauss}} = \frac{1}{\eta^2} \left( \mathbf{x} - \bigodot_{i=1}^K \mathbf{x}_i \right) \odot \bigodot_{i \ne j} \mathbf{x}_i.
\end{align}

\paragraph{Similarity energy}  
Alternatively, we can define a similarity-based energy
\begin{align}
    \mathcal{E}_{\text{sim}}(\mathbf{x}_{1:K}; \mathbf{x}) = \mathbf{x}^\top \left( \bigodot_{i=1}^K \mathbf{x}_i \right)-\frac{\lambda}{2}\sum_{i=1}^K\|\mathbf{x}_i\|^2
\end{align}
which encourages the estimated bound factor to align with the observation in direction while penalizing large magnitudes. The corresponding gradient is
\begin{align}\label{eq:sim-energy}
    \nabla_{\mathbf{x}_j} \mathcal{E}_{\text{sim}} = \mathbf{x} \odot \bigodot_{i \ne j} \mathbf{x}_i-\lambda\mathbf{x}_j
\end{align}
Although $\exp(\mathcal{E}_{\text{sim}})$ does not define a proper likelihood over $\mathbb{R}^D$ as the normalization constant diverges, the gradient is well-defined and can be used directly for guidance.

Both energy functions are evaluated at the denoised estimates $\hat{\mathbf{x}}_i^0(t)$ via the DPS approximation in Eq.~\ref{eq:dps-approx}. We use the noisy estimate at time $t$ the compute the regularization term in Eq.~\ref{eq:sim-energy}.

\subsection{Iterative Sampling}\label{sec:it-sampling}
To refine factor estimates, we employ an iterative sampling strategy. Let $T$ be the number of discretized diffusion steps, $R$ the number of iterative restarts, and $\rho\in[0,1]$ be the restart ratio. Let $\tau=\frac{\lfloor \rho T \rfloor}{T}$ be the restart time. 

To perform iterative sampling, we first initialize estimates $\tilde{\mathbf{x}}_j^0=\frac{1}{n}\sum_{\mathbf{x}\in X_j}\mathbf{x}$. We then sample $\tilde{\mathbf{x}}_j^\tau\sim p(\mathbf{x}_j^\tau | \mathbf{x}_j^0=\frac{1}{n}\sum_{\mathbf{x}\in X_j}\mathbf{x})$ and simulate the reverse diffusion process from $t=\tau$ to $t=0$. We use the result of this sampling procedure for the next sampling iteration, repeating the process $R$ times. Thus, the total number of iterations of this process is $\kappa =R \lfloor \rho T \rfloor$. Note that \textit{the computational complexity of a single iteration is the same as that of a resonator network.}
\section{Results}
\paragraph{Dataset Generation}
We generate codebooks $X_j\in\{-1,1\}^{D\times n}$, $j=1,\dots,K$, where each entry is sampled from $\mathcal{U}(\{-1,1\})$. From each codebook, we select a random codebook vector $\mathbf{x}_j\in X_j$ and compute the bound vector $\mathbf{x}=\mathbf{x}_1\odot\cdots\odot\mathbf{x}_K$.

To evaluate the models, we compute their decomposition accuracy $c/K$, where $c$ is the number of correct factors. We report the mean decomposition accuracy over 200 trials.

\begin{table*}[!t]
\centering
\caption{Decomposition accuracy at different noise levels $\sigma=\sqrt{m-1}$. We fix the codebook dimension to be $D=1000$ and a search space size of $10^4$ with 100 iterations.}
\label{tab:acc-vs-noise}
\begin{tabular}{c l ccc ccc ccc}
\toprule
 & \textbf{Model} & \multicolumn{3}{c}{$K=2$} & \multicolumn{3}{c}{$K=3$} & \multicolumn{3}{c}{$K=4$} \\
\cmidrule(lr){3-5} \cmidrule(lr){6-8} \cmidrule(lr){9-11}
 & & $m=2$ & $m=4$ & $m=8$ & $m=2$ & $m=4$ & $m=8$ & $m=2$ & $m=4$ & $m=8$ \\
\midrule
& \textsc{Resonator}         & \textbf{1.000} & 0.957 & 0.565 & \textbf{0.995} & 0.795 & 0.256 & 0.596 & 0.336 & 0.109 \\
& \textsc{Attn. Resonator}   & 0.885 & 0.533 & 0.250 & 0.735 & 0.351 & 0.107 & 0.608 & 0.310 & 0.136 \\
& \textsc{ALS}               & 0.977 & 0.625 & 0.190 & 0.830 & 0.523 & 0.285 & 0.580 & 0.368 & 0.203 \\
\midrule
& \textsc{Gaussian}          & 0.340 & 0.530 & 0.370 & 0.047 & 0.065 & 0.055 & 0.097 & 0.099 & 0.131 \\
& \textsc{Latent Gaussian}   & \textbf{1.000} & 0.955 & \textbf{0.690} & \textbf{0.995} & 0.880 & \textbf{0.510} & 0.827 & \textbf{0.674} & \textbf{0.454} \\
& \textsc{Similarity}        & \textbf{1.000} & 0.990 & 0.675 & \textbf{0.995} & \textbf{0.885} & 0.467 & \textbf{0.904} & 0.647 & 0.445 \\
& \textsc{Latent Similarity} & \textbf{1.000} & \textbf{0.993} & 0.655 & 0.975 & 0.862 & 0.427 & 0.746 & 0.608 & 0.384 \\
\bottomrule
\end{tabular}
\end{table*}

\subsection{Baseline Comparisons}
\paragraph{Baselines}
We consider three different baseline models for decomposition: (1) resonator networks \citep{fradyResonatorNetworksEfficient2020}, the current state of the art; (2) attention-based resonator networks \citep{yeungSelfAttentionBasedSemantic2024}; and (3) alternating least-squares (ALS) \citep{tensor-decomp,kentResonatorNetworks22020}. See section \ref{sec:baselines} in the appendix for more information on these decomposition methods.

\paragraph{Model Variants}
As noted in sections \ref{sec:decomp-inv} and \ref{sec:joint-inf} respectively, there are two main design choices: (1) whether the diffusion occurs in the codebook vector space or latent space; and (2) whether we use a Gaussian- or similarity-based energy function for coupled guidance. These two design choices result in four model variants, which we denote as \textsc{Gaussian}, \textsc{Latent Gaussian}, \textsc{Similarity}, and \textsc{Latent Similarity}. We simulate the reverse diffusion process using the probability-flow ODE.

Given a computational budget of $N=100$ iterations, we find the optimal hyperparameters for each model variant using Optuna \citep{optuna_2019}. We explain this process in more detail in section \ref{sec:impl-details} of the appendix.

\paragraph{Decomposition accuracy} We evaluate all models on different search space sizes $n^K$ and different number of codebooks $K$, visualized in figure \ref{fig:acc-vs-search-space}. \textsc{Similarity} and \textsc{Latent Similarity} significantly outperform all other models. Here, we choose a codebook vector dimension of $D=1000$ and all models have a computational budget of 100 iterations.

We define the capacity of a model to be the largest search space for which the model achieves an accuracy over $\alpha=95\%$. \textsc{Similarity} achieves a capacity $2\times$ that of resonator networks, the current state of the art, over different values of $K$. Notably, \textsc{Latent Similarity} has a capacity $5.38\times$ that of resonator networks for $K=5$.

\textsc{Gaussian} performs approximately similarly to the the baselines, while \textsc{Latent Gaussian} performs similar to the similarity model variants for $K=3,4$ and closer to \textsc{Gaussian} for $K=5$. 

\paragraph{Decomposing superposed representations}
It is common for binding-based representations to consist of a superposition of bound vectors $\mathbf{x}=\sum_{i=1}^m\mathbf{x}_1^i\odot\dots\odot\mathbf{x}_K^i$. For example, scenes can be described as superpositions of the objects present, each with different attributes. More concretely, an image of a basket of apples could be represented as $\sum_{j=1}^m\mathbf{size}_j\odot \mathbf{color}_j\odot \mathbf{position}_j$, where $j$ indexes each apple. 

To recover the attributes of each object, we can apply a decomposition method to the superposed representation and have it converge onto one of the bound vectors, then subtract the result from the superposed representation. Repeating this process can recover the attributes of all objects. Thus, it is important to consider the probability that the result converges to the factors of one of the bound vectors in the superposition. In the worst case, we can evaluate this capability by computing the decomposition accuracy of the model when Gaussian noise is added to the bound representation $\mathbf{x}=\mathbf{x}_1\odot\dots\odot\mathbf{x}_K+\sigma\epsilon$, $\epsilon\sim\mathcal{N}(0,I)$, where $\sigma=\sqrt{m-1}$.

\begin{figure}
    \centering
    \includegraphics[width=0.6\linewidth]{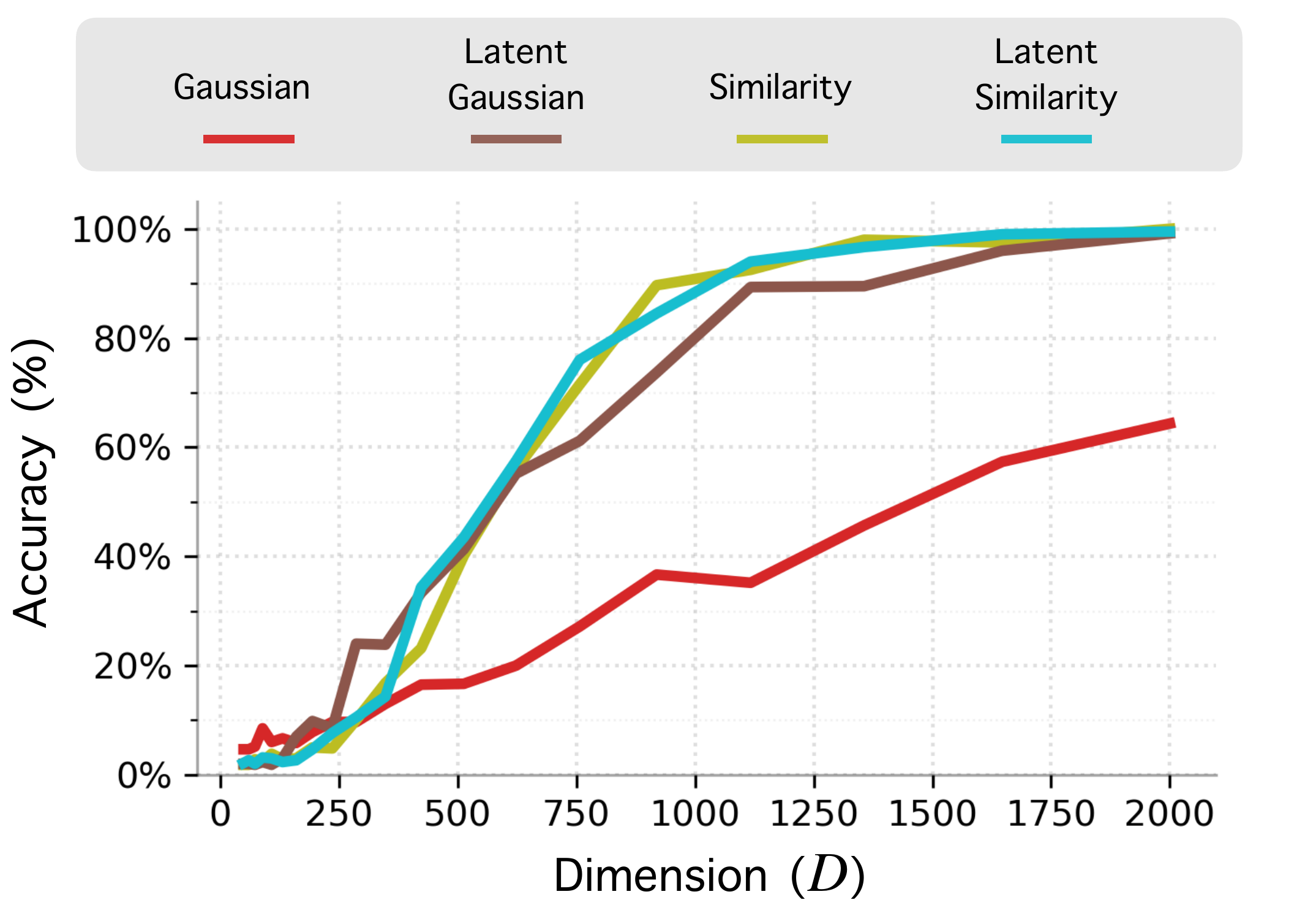}
    \caption{Decomposition accuracy as codebook vector dimension $D$ varies, for $K=3,n=50$.}
    \label{fig:acc-vs-d}
\end{figure}

Table \ref{tab:acc-vs-noise} reports the effect of added noise to decomposition accuracy, with the search space size fixed to $10^4$. We find 
\textsc{Latent Gaussian} and \textsc{Similarity} models to significantly outperform baseline models, especially for larger $K$ and $m$. For example, \textsc{Similarity} and \textsc{Latent Gaussian} achieve an 11\% and 22.1\% increase in accuracy over resonator networks for $(K,m)=(3,4)$ and $K=2$, respectively.


\subsection{Ablations}
\paragraph{Effect of codebook vector dimension} 
As we sample codebook elements from $\{-1,1\}$ uniformly, codebook vectors are quasi-orthogonal. Increasing $D$ increases their degree of quasi-orthogonality. As shown in figure \ref{fig:acc-vs-d}, increasing dimension reliably increases decomposition accuracy, with marginal improvement as the accuracy approaches 100\%.

\begin{figure}
    \centering
    \includegraphics[width=0.6\linewidth]{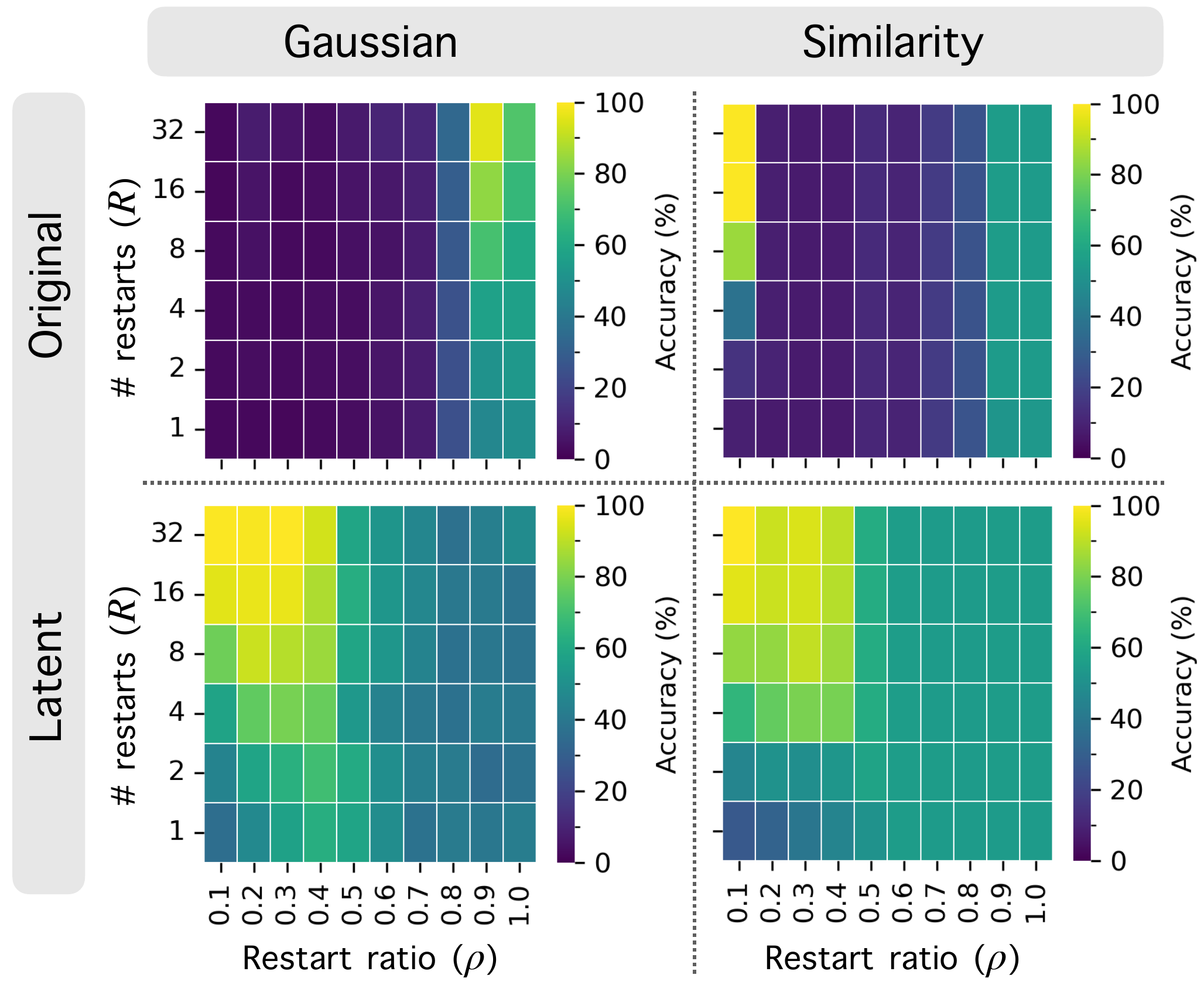}
    \caption{Decomposition accuracy when varying restart ratio $\rho$ and number of restarts $R$ for $D=1000$, $K=3$, and $n=40$.}
    \label{fig:acc-vs-it}
\end{figure}

\paragraph{Effect of iterations}
Three factors determine the number of iterations: the restart ratio $\rho$, number of diffusion steps $T$, and number of restarts $R$.

Fixing the other hyperparameter settings for each model variant, we vary $\rho$ and $R$ to evaluate its effect on decomposition accuracy, as shown in figure \ref{fig:acc-vs-it}. We find that in general, higher $R$ leads to higher accuracy. This is not surprising since if each restart has an independent probability $p$ of being correct, then the entire iterative sampling process has a probability of $1-p^R$ of being correct after $R$ restarts, which converges to $1$ as $R\to\infty$.

We observe that the optimal $\rho$ for \textsc{Gaussian} is close to 1, while the optimal value for the other model variants is closer to 0. In addition, for non-latent model variants, values of $\rho$ that yield high accuracy are tightly concentrated around the optimal value, whereas it is more dispersed for latent model variants. 

\begin{figure}
    \centering
    \includegraphics[width=0.6\linewidth]{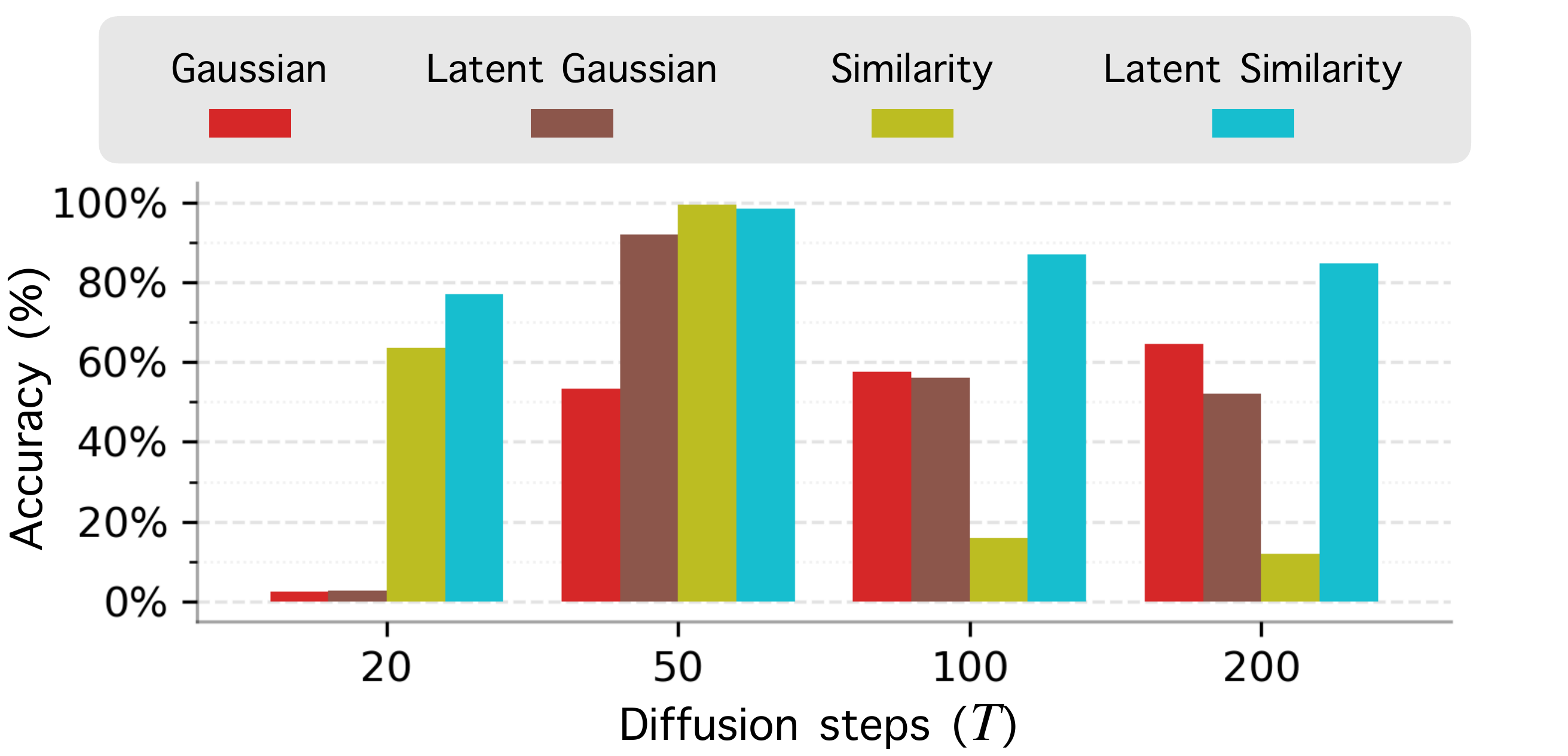}
    \caption{Decomposition accuracy when varying the number of discretized diffusion steps for $D=1000$, $K=3$, and $n=40$.}
    \label{fig:acc-vs-T}
\end{figure}

Holding other hyperparameters constant, we also vary $T$ to evaluate its effect of decomposition accuracy. For general applications of diffusion, a larger $T$ results in better performance as it results in a more accurate simulation of the reverse process. However, this trend does not hold for our model, as no clear trend can be seen in figure \ref{fig:acc-vs-T}. 

Choices of $\rho,R,T$ enable us to interpolate between continuous and discrete updates, which correspond to $\rho=R=1$ with large $T$, and large $R$ with small $\rho,T$, respectively. This suggests a continuum between associative memories like Hopfield networks, with highly discretized rules, and diffusion models, which rely on smaller discretization steps. The results in figure \ref{fig:acc-vs-T} suggest that the optimal choices of these parameters correspond to models which lie on this continuum.

\paragraph{ODE vs SDE}
The results reported above use the probability-flow ODE to simulate the reverse process. We compute the decomposition accuracy for \textsc{Gaussian}, \textsc{Latent Gaussian}, \textsc{Similarity}, and \textsc{Latent Similarity} models with the same hyperparameters, but simulate the reverse process using the reverse SDE instead of probability-flow ODE.

\begin{figure}
    \centering
    \includegraphics[width=0.6\linewidth]{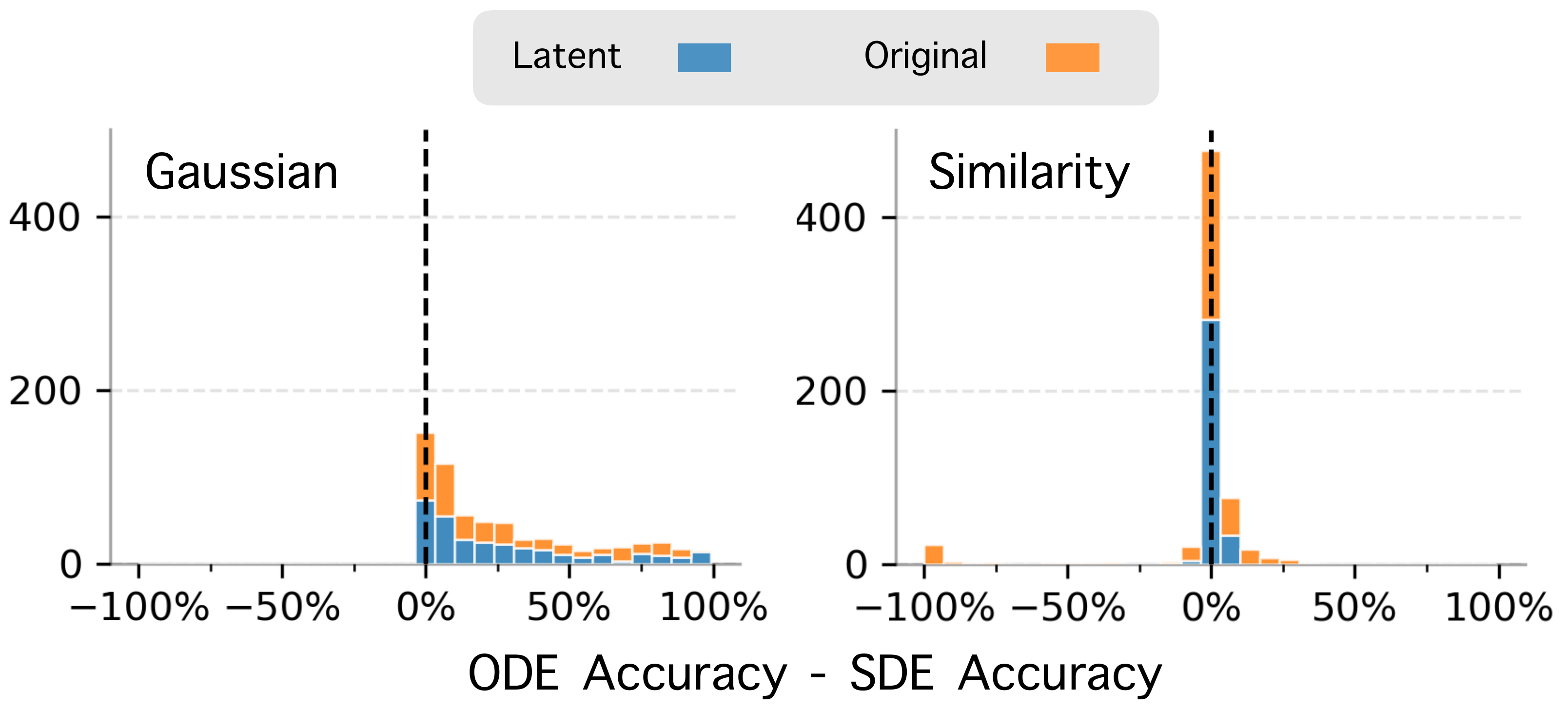}
    \caption{Histogram of differences in accuracy between ODE- and SDE-based simulation of the reverse diffusion process.}
    \label{fig:ode-vs-sde}
\end{figure}

Figure \ref{fig:ode-vs-sde} plots a histogram of differences in accuracy between ODE- and SDE-based simulation of the reverse diffusion process. We find that there is a difference in behavior determined by the kind of energy used for guidance. While guidance using similarity energy yields no significant difference between ODE and SDE models, we find that the ODE variant outperforms its SDE counterpart when using Gaussian energy for guidance. This result holds for both latent and non-latent variants.

\section{Discussion}
\subsection{Relation to Resonator Networks}\label{sec:relation-to-rn}
Consider the case in which we use a Gaussian energy where $\eta^2=0$. This imposes a constraint where $\mathbf{x}=\mathbf{x}_1^0\odot\cdots\odot \mathbf{x}_K^0$. Assuming $\mathbf{x}_j^0$ are bipolar codebook vectors,
\begin{align}
    \mathbf{x}_j^0=\mathbf{x}\odot \bigodot_{i \ne j} \mathbf{x}_i^0
\end{align}
Approximating $\mathbf{x}_j^0$ with Tweedie's estimate, we get
\begin{align}\label{eq:tweedie-approx}
    \hat{\mathbf{x}}_j^0(t)\approx\mathbf{x}\odot \bigodot_{i \ne j} \hat{\mathbf{x}}_i^0(t)
\end{align}
Under the assumption that $\sigma_0^2=0$ and that codebook vectors are normalized, remark \ref{rem:softmax} notes that $\hat{\mathbf{x}}_i^0(t)$ is a single application of the softmax update rule of a modern Hopfield network. 

Now, following a deterministic version of the iterative sampling procedure described in section \ref{sec:it-sampling}, if we replace the simulation of the reverse ODE at each sampling iteration with the approximation in Eq.~\ref{eq:tweedie-approx}, corresponding to particular selections of $\rho$, $T$, and $R$, we recover an update rule very similar to the resonator update rule given in Eq.~\ref{eq:resonator-update} but with a softmax-based update rule as in \citep{yeungSelfAttentionBasedSemantic2024}. Thus, \textit{one can view our proposed method as a continuous generalization of the softmax-based resonator network.}

In addition, this perspective suggests an analogy between resonator networks and our proposed method. As noted in section \ref{sec:resonator}, resonator networks are $K$ coupled Hopfield networks. Section \ref{sec:diffusion} notes the similarity between Hopfield networks and diffusion models. These two observations motivate a design in which $K$ diffusion models are coupled together for semantic decomposition. Thus, \textit{our model can be viewed as the diffusion model equivalent of resonator networks.}

\subsection{Relation to Compositional Diffusion}
Compositional diffusion involves combining preexisting diffusion models to create a new model without training. This idea has been explored in the diffusion literature through the introduction of compositional operators on diffusion models, such as conjunction and disjunction, and more sophisticated sampling schemes \citep{liuCompositionalVisualGeneration,duReduceReuse}. These operators produce a composite model whose corresponding distribution is the product and mixture of its constituent distributions, respectively.

Our method can also be considered a form of compositional diffusion; it takes preexisting diffusion models modeling individual codebooks and combines them to produce a model that operates on the product space of these codebooks. Specifically, let $\mathcal{X}_j$ be the space the $j$-th diffusion model operates on. Then the composite model operates on $f(\mathcal{X}_1\times\dots\times\mathcal{X}_K)$, where $f(\mathbf{x_1},\dots,\mathbf{x}_K)=\bigodot_{j=1}^K \mathbf{x}_j$. The resulting distribution modeled by the composite diffusion model is induced by the guidance term.

\subsection{Relation to Blind Inverse Problems}
A blind inverse problem is an inverse problem where the form of the degradation function is known but not its hyperparameters \citep{darasSurveyDiffusionModels2024}. One can characterize the decomposition problem as a blind inverse problem as follows:
\begin{align}
    \mathbf{x}=A\mathbf{x}_j,\quad A=\bigodot_{i\neq j} \mathbf{x}_i
\end{align}
This characterization is symmetric, so for any choice of $j$, $\{\mathbf{x}_i\}_{i\neq j}$ form the hyperparameters of its corresponding blind inverse problem. BlindDPS \citep{chungParallelDiffusionModels2023} is an extension of DPS \citep{chungDiffusionPosteriorSampling2024} that uses parallel diffusion process to model the unknown hyperparameters and the DPS estimate for the conditional guidance term. This approach resembles our method, but we make additional design choices motivated by the view of diffusion models as an associative memory, such as iterative sampling and higher levels of discretization.
\section{Conclusion}
We presented a framework for semantic decomposition that couples diffusion processes through joint guidance. We construct analytic diffusion priors and define an energy function that aligns factor estimates with a given bound representation. We use an iterative sampling scheme to refine factor estimates without changing the asymptotic per-iteration cost, leading to improved capacity and robustness over existing baselines. These results indicate that diffusion models can act not only as powerful image priors but also be coupled to solve compositional combinatorial tasks such as semantic decomposition.

\section*{Acknowledgments}
This work was supported in part by the DARPA Young Faculty Award, the National Science Foundation (NSF) under Grants \#2127780, \#2319198, \#2321840, \#2312517, and \#2235472, \#2431561, the Semiconductor Research Corporation (SRC), the Office of Naval Research through the Young Investigator Program Award, and Grants \#N00014-21-1-2225 and \#N00014-22-1-2067, Army Research Office Grant \#W911NF2410360. Additionally, support was provided by the Air Force Office of Scientific Research under Award \#FA9550-22-1-0253, along with generous gifts from Xilinx and Cisco.


\appendix
\section{Coupled Latent Inference for Semantic Decomposition}\label{sec:latent-decomp}
Following the characterization of the decomposition problem in Eq.~\ref{eq:latent-decomp} yields a solution based on latent diffusion. The methodology is similar to that described above. Here, we highlight the main differences.

\paragraph{Score Function} The score function is
\begin{align}\label{eq:uncond-score-z}
    \nabla_{\mathbf{z}_j^t} \log p_j(\mathbf{z}_j^t) &= \frac{1}{\omega^2_t}\sum_{\mathbf{z}\in I_n}\gamma_\mathbf{z}(\mathbf{z}_j^t,t)(\alpha_t\mathbf{z}-\mathbf{z}_j^t)\\
    \text{where}\quad\gamma_\mathbf{z}(\mathbf{z}_j^t,t) &= \frac{\mathcal{N}(\mathbf{z}_j^t;\alpha_t\mathbf{z},\omega_t^2I)}{\sum_{\mathbf{z}'\in I_n} \mathcal{N}(\mathbf{z}_j^t;\alpha_t\mathbf{z}',\omega_t^2I)}
\end{align}

\paragraph{Gaussian energy} The Gaussian energy is
\begin{align}
    \mathcal{E}_{\text{Gauss}}(\mathbf{z}_{1:K}; \mathbf{x}) = -\frac{1}{2\eta^2} \left\| \mathbf{x} - \bigodot_{i=1}^K X_i\mathbf{z}_i \right\|^2
\end{align}
resulting in the gradient
\begin{align}
    \nabla_{\mathbf{x}_j} \mathcal{E}_{\text{Gauss}} = \frac{1}{\eta^2} X_j^\top\left(\left( \mathbf{x} - \bigodot_{i=1}^K X_i\mathbf{z}_i \right) \odot \bigodot_{i \ne j} X_i\mathbf{z}_i\right).
\end{align}

\paragraph{Similarity energy}  
The similarity-based energy is
\begin{align}
    \mathcal{E}_{\text{sim}}(\mathbf{z}_{1:K}; \mathbf{x}) = \mathbf{x}^\top \left( \bigodot_{i=1}^K X_i\mathbf{z}_i \right)-\frac{\lambda}{2}\sum_{i=1}^K\|\mathbf{x}_i\|^2
\end{align}
resulting in the gradient
\begin{align}
    \nabla_{\mathbf{x}_j} \mathcal{E}_{\text{sim}} = X_j^\top\left(\mathbf{x} \odot \bigodot_{i \ne j} X_i\mathbf{z}_i\right)-\lambda\mathbf{x}_j
\end{align}

\section{Other Decomposition Methods}\label{sec:baselines}
\paragraph{Attention Resonator Network}
The attention resonator network \citep{yeungSelfAttentionBasedSemantic2024} uses an update rule similar to the resonator network with an additional softmax, motivated by the design of modern Hopfield networks \citep{ramsauerHopfieldNetworksAll2021}:
\begin{align}
\hat{\mathbf{x}}^{(j)}_{t+1} &= X_j \mathrm{softmax}(\beta X_j^\top(\mathbf{x}\odot \mathbf{r}_t^{(j)}))
\end{align}
We use $\beta=250$ in the evaluations.

\paragraph{Alternating Least Squares} Alternating least squares (ALS) is a commonly used method for tensor decomposition \citep{tensor-decomp,kentResonatorNetworks22020}. It iterates through each factor and solves a least-square problem $\|\mathbf{x}-\bigodot_{j=1}^K X_j\mathbf{z}_j\|^2$ to obtain the new factor estimate. This procedure is repeated $N$ times, where $N$ is the number of iterations.

Solving for $\mathbf{z}_j$, we have to compute $\mathbf{z}_j=A^\dagger\mathbf{x}$, where $A^\dagger$ is the pseudoinverse of $A=X_j\prod_{i\neq j}\mathrm{diag}(X_i\mathbf{z}_i)$. Computing the pseudoinverse is computationally prohibitive for large $D$.

\section{Implementation Details}\label{sec:impl-details}
\paragraph{Codebook Generation}
We generate codebooks $X_j\in\{-1,1\}^{D\times n}$, $j=1,\dots,K$, where each entry is sampled from $\mathcal{U}(\{-1,1\})$. Codebook entries are scaled to have a magnitude of 1000.

\paragraph{Noise Schedule}
We use a linear noise schedule, where $\alpha_t,\beta_t$ in Eq.~\ref{eq:marginal} is determined by $\alpha_t=\sqrt{\bar{a}_t},\beta_t^2=1-\bar{a}_t$, where
\begin{align}
    \bar{a}_t&=\prod_{s=1}^t a_s\\
    a_t&=1-b_t\\
    b_t&=b_\mathrm{min}+\frac{t-1}{T-1}(b_\mathrm{max}-b_\mathrm{min})
\end{align}
with $b_\mathrm{min}=0.1,b_\mathrm{max}=20$.

\paragraph{Guidance Strength Schedule}
We are given a base guidance strength $\eta$, while is subsequently scaled depending on the schedule. These scheudles include: (1) constant, (2) linear, (3) SNR, and (4) $\sigma$. 

The constant schedule is self explanatory. The linear schedule linearly interpolates the scaling factor from $\eta_\mathrm{min}$ to $\eta_\mathrm{\max}$ from $t=T,\dots,1$. The $\sigma$ schedule scales the guidance strength by $\beta_t$, while the SNR schedule scales it by $\beta_t/\alpha_t$.

\paragraph{Guidance Clipping}
To deal with unstable gradients, we ensure the magnitude of the guidance vector does not exceed some scalar multiple of the unconditional score. If the guidance vector exceeded this maximum magnitude, it is rescaled to have the same magnitude it.

\paragraph{Iterative Sampling}
To perform iterative sampling, we first initialize estimates $\tilde{\mathbf{x}}_j^0=\frac{1}{n}\sum_{\mathbf{x}\in X_j}\mathbf{x}$. We then sample $\tilde{\mathbf{x}}_j^\tau\sim p(\mathbf{x}_j^\tau | \mathbf{x}_j^0=\frac{1}{n}\sum_{\mathbf{x}\in X_j}\mathbf{x})$ and simulate the reverse diffusion process from $t=\tau$ to $t=0$. We use the result of this sampling procedure for the next sampling iteration, repeating the process $R$ times.

The process of sampling $\tilde{\mathbf{x}}_j^\tau\sim p(\mathbf{x}_j^\tau | \mathbf{x}_j^0=\frac{1}{n}\sum_{\mathbf{x}\in X_j}\mathbf{x})$ can be stochastic or deterministic (by predicting the mean). Throughout this work, we use the deterministic variant. Adding stochasticity does not seem to have significant impact on performance. We leave further exploration for future work.

\paragraph{Hyperparameters Optimization}
We use Optuna \citep{optuna_2019} to optimize the following hyperparameters:
\begin{enumerate}
    \item $\eta$: inverse guidance strength
    \item $\lambda$ schedule: chosen between \texttt{constant}, \texttt{linear}, \texttt{snr}, \texttt{sigma}
    \item \texttt{cond\_clip\_ratio}: ensures the magnitude of the guidance vector does not exceed \texttt{cond\_clip\_ratio} times that of the magnitude of the unconditional score
    \item $\sigma_0$: the noise at $t=0$ which determines the strength of the attractors at each codebook vector
    \item \texttt{softmax\_temp}: an additional temperature parameter used in the computation of Tweedie's estimate; this is effectively equivalent to scaling the codebook vectors
    \item $\lambda$: regularization strength
    \item $T$: discretization steps
    \item $R$: number of restarts
    \item $\rho$: restart ratio
\end{enumerate}
The optimization is done on a validation dataset, with the goal of maximizing the decomposition accuracy for $D=1000,K=3,n=50$. The hyperparameter optimization for each model variant is done separately.

\subsection{Optimal Configurations}
\paragraph{Gaussian}
The best \textsc{Gaussian} configuration uses an inverse guidance strength of $\eta = 0.0171$ with a $\sigma$ guidance schedule, i.e., $\lambda$ schedule = \texttt{sigma}. We set \texttt{cond\_clip\_ratio} $= 2.81$ to bound the conditional guidance relative to the unconditional score, use a base noise level of $\sigma_0 = 0.110$, a softmax temperature of \texttt{softmax\_temp} $= 13.63$ , and a regularization strength of $\lambda = 88.47$. The SDE is discretized with $T = 50$ steps, and we employ $R = 2$ restarts with restart ratio $\rho = 1$.

\paragraph{Latent Gaussian}
The best \textsc{Latent Gaussian} configuration uses an inverse guidance strength of $\eta = 0.0357$ with a constant guidance schedule, i.e., $\lambda$ schedule = \texttt{constant}. We set \texttt{cond\_clip\_ratio} $= 5.44$ to bound the conditional guidance relative to the unconditional score, use a base noise level of $\sigma_0 = 0.314$, a softmax temperature of \texttt{softmax\_temp} $= 7.04$ , and a regularization strength of $\lambda = 41.86$. The SDE is discretized with $T = 50$ steps, and we employ $R = 14$ restarts with restart ratio $\rho \approx 0.143$.

\paragraph{Similarity}
The best \textsc{Similarity} configuration uses an inverse guidance strength of $\eta = 0.106$ with a $\sigma$ guidance schedule, i.e., $\lambda$ schedule = \texttt{sigma}. We set \texttt{cond\_clip\_ratio} $= 1.44$ to bound the conditional guidance relative to the unconditional score, use a base noise level of $\sigma_0 = 0.0408$, a softmax temperature of \texttt{softmax\_temp} $= 4.22$, and a regularization strength of $\lambda = 82.8$. The SDE is discretized with $T = 50$ steps, and we employ $R = 20$ restarts with restart ratio $\rho = 0.1$.

\paragraph{Latent Similarity}
The best \textsc{Latent Similarity} configuration uses an inverse guidance strength of $\eta = 0.0167$ with a $\sigma$ guidance schedule, i.e., $\lambda$ schedule = \texttt{sigma}. We set \texttt{cond\_clip\_ratio} $= 2.40$ to bound the conditional guidance relative to the unconditional score, use a base noise level of $\sigma_0 = 0.0317$, a softmax temperature of \texttt{softmax\_temp} $= 1.11$, and a regularization strength of $\lambda = 36.33$. The SDE is discretized with $T = 50$ steps, and we employ $R = 20$ restarts with restart ratio $\rho = 0.1$.

\end{document}